\documentclass[acmsmall,screen,nonacm]{acmart}
\AtBeginDocument{%
  }

\setcopyright{acmlicensed}
\copyrightyear{2027}
\acmYear{2027}
\acmDOI{10.1145/XXXXXXX.XXXXXXX}
\acmConference[FSE '27]{ACM International Conference on the Foundations of Software Engineering}{July 12--16, 2027}{Shenzhen, China}
\acmISBN{978-1-4503-XXXX-X/2018/06}
\usepackage{xcolor}
\usepackage{booktabs,makecell,tabularx,array}
\usepackage{graphicx,multirow} 
\usepackage{subcaption}
\usepackage{pifont}

\usepackage{tcolorbox}
\tcbuselibrary{skins}
\newtcolorbox{findingbox}{
    enhanced,
    colback=gray!12,    
    colframe=white,     
    sharp corners,
    boxrule=0pt,
    top=1.2mm,
    bottom=1.2mm,
    left=1.8mm,
    right=1.8mm,
    before skip=4pt,    
    after skip=4pt     
}

\usepackage{booktabs}   
\usepackage{makecell}   
\usepackage{multirow}   
\usepackage{graphicx}   
\usepackage{xcolor}     
\usepackage{array}      
\usepackage{tabularx}   

\usepackage{tikz}
\usetikzlibrary{positioning,arrows.meta,fit,backgrounds,calc}

\begin{document}

\title{Why Static Skill Inconsistencies Do Not Equate to Runtime Risks: A Behavioral Analysis of LLM Agent Execution}

\author{Fengrui Fan}
\email{fengruifan@hust.edu.cn}
\orcid{0009-0008-0828-3769}
\affiliation{%
  \institution{Huazhong University of Science and technology}
  \city{Wuhan City}
  \state{Hubei Province}
  \country{China}}


\author{Huahao Wang}
\email{wanghuahao20040112@gmail.com}
\orcid{0009-0000-2502-7697}
\affiliation{%
  \institution{Huazhong University of Science and technology}
  \city{Wuhan City}
  \state{Hubei Province}
  \country{China}}

\author{JingYi Fang}
\email{fangjingyi@hust.edu.cn}
\orcid{0009-0007-4898-4124}
\affiliation{%
  \institution{Huazhong University of Science and technology}
  \city{Wuhan City}
  \state{Hubei Province}
  \country{China}}

\author{Yutao Hu}
\email{yutaohu@hust.edu.cn}
\orcid{0009-0007-4898-4124}
\correspondingauthor
\affiliation{%
  \institution{Huazhong University of Science and technology}
  \city{Wuhan City}
  \state{Hubei Province}
  \country{China}}

\author{Yueming Wu}
\email{yuemingwu@hust.edu.cn}
\orcid{0009-0007-4898-4124}
\affiliation{%
  \institution{Huazhong University of Science and technology}
  \city{Wuhan City}
  \state{Hubei Province}
  \country{China}}

\author{Deqing Zou}
\email{deqingzou@hust.edu.cn}
\orcid{0000-0001-8534-5048}

\affiliation{%
  \institution{Huazhong University of Science and technology}
  \city{Wuhan City}
  \state{Hubei Province}
  \country{China}}

\renewcommand{\shortauthors}{Trovato et al.}


\begin{abstract}
LLM agent frameworks heavily rely on external skill assets to standardize tool usage, encode operational workflows, and supply executable runtime resources. Despite their widespread adoption, multi-component skill packages frequently suffer from cross-artifact inconsistencies, where documentation, scripts, and configuration files specify conflicting behavioral constraints. Although static auditing techniques can uncover such contractual discrepancies, static defect volume cannot reliably reflect real-world agent performance degradation. This gap leaves practitioners without rigorous evidence to assess the practical risk of flawed agent skills.

This paper presents a systematic empirical study to demystify the runtime implications of skill inconsistencies. We conduct large-scale auditing over 575 community and benchmark skills, yielding 1,182 manually validated contract inconsistencies. We further build a controlled evaluation pipeline with 49 executable agent tasks and three representative LLMs across NoSkill, OriginalSkill, and FixedSkill experimental settings. Through fine-grained behavioral analysis of 441 paired execution trajectories, we quantify the actual performance impacts of resolved defects.

Our empirical results reveal that contractual conflicts are pervasive in practical skill ecosystems, impacting over 64\% of all examined skills. Skill repairing consistently improves task success rates, yet the performance benefits exhibit strong model heterogeneity, ranging from substantial improvement to negligible gain. Further trajectory dissection indicates that nearly all execution instances expose inconsistent skill content, which rules out exposure frequency as the source of performance variance. Instead, cross-model behavioral differences after inconsistency exposure dominate final task outcomes. We find that weaker models frequently follow erroneous skill specifications and propagate defects into final results, while stronger models actively mitigate runtime errors. Our findings demonstrate that the practical risk of flawed skills is jointly determined by execution reachability and agent post-execution behavior, rather than static inconsistency quantity alone.
\end{abstract}





\maketitle

\section{Introduction}

\noindent\textbf{Why Agent Skills Matter.}
LLM agents increasingly rely on reusable components to acquire procedural knowledge, tool-use capabilities, and domain-specific workflows. 
Agent Skills are an important artifact in this emerging ecosystem \cite{wang2023voyager,jiang2026sok}. 
Unlike one-off prompts, a Skill is typically distributed as a reusable package containing documentation, metadata, scripts, configuration files, and other resources \cite{anthropic-agent-skills-docs,anthropic-agent-skills-news}. 
Together, these artifacts specify how the Skill should be discovered, interpreted, and executed.
As the public Skill ecosystem rapidly expands, the same Skill may be reused across different tasks and agents, allowing quality issues to affect multiple downstream executions.

This multi-artifact structure also introduces a new reliability challenge. 
Because individual artifacts can evolve independently, they may make incompatible claims about the same functionality, interface, or runtime requirement. 
For example, documentation may promise a capability that the implementation does not provide, or different documents may specify conflicting dependency versions. 
We refer to such cases as \emph{Skill contract inconsistencies}. 
Unlike conventional specification-implementation mismatches, these inconsistencies may arise between a Skill and its environment, across documentation artifacts, or between documentation and code.

\noindent\textbf{Do Static Inconsistencies Matter at Runtime?}
Detecting a static inconsistency does not imply that it will cause runtime harm. 
Static inspection can establish that two artifacts conflict, but it cannot determine whether the conflict will enter actual execution or whether it will ultimately affect task outcomes or execution cost. 
A static finding therefore indicates a potential problem, not its actual runtime risk.

This distinction has direct practical consequences. 
If all inconsistencies are treated equally, limited auditing and repair effort cannot be prioritized toward findings that actually affect execution, while low-impact findings may obscure more consequential ones. 
Skill quality assessment therefore needs to move beyond finding detection toward runtime risk prioritization.

Existing work has examined Skill structure \cite{zhou2026comprehensive}, documentation quality \cite{li2026skillsbench}, security \cite{guo2026malskillbench, jiang2026harmfulskillbench}, and artifact-level defects \cite{zhang2026keeps,meng2026skillconsist}. 
However, there is still limited evidence on whether such static findings enter agent execution, whether repairing them changes runtime outcomes, and how agents respond after exposure. 
We therefore lack a systematic understanding of how \emph{static defects} translate into \emph{runtime consequences}: how prevalent Skill contract inconsistencies are, how much they affect agent execution, and why these effects arise.

\noindent\textbf{Our Study and Key Findings.}
To answer these questions, we conduct an empirical study of Skill contract inconsistencies. First, we collect 575 Skills from GitHub and SkillBench \cite{li2026skillsbench} and develop a taxonomy spanning three contract boundaries: Environment, Inter Document, and Document Code. 
Through automated auditing and manual validation, we confirm 1,182 inconsistencies and characterize their prevalence, types, and distribution. 
Second, we construct 49 verifiable tasks from real-world Skills and evaluate three models under three conditions: \textsc{NoSkill}, \textsc{OriginalSkill}, and \textsc{FixedSkill}. 
The comparison between \textsc{OriginalSkill} and \textsc{FixedSkill} estimates the runtime effect of the targeted inconsistency, while \textsc{NoSkill} separates this effect from the general benefit of using a Skill. Finally, we analyze paired execution trajectories to trace whether inconsistencies enter execution and how different responses relate to final outcomes.

Our study yields four main findings. 
\ding{182} Skill contract inconsistencies are widespread and unevenly distributed. At least one inconsistency appears in 67.4\% of GitHub Skills and 64.4\% of SkillBench Skills. 
In SkillBench, Document Code inconsistencies account for 56.2\% of all findings, with \texttt{over\_promised} alone accounting for 45.6\%.
\ding{183} Fixing inconsistencies improves average task effectiveness for all three models, but the magnitude varies substantially. Claude Haiku 4.5 \cite{anthropic-haiku45}, Qwen3.6-35B \cite{qwen36-35b,omlx-qwen36}, and DeepSeek-V4-Flash \cite{deepseek-v4,continuum-v4flash,macaron-v4bench} improve by 0.136, 0.085, and 0.023 score points, respectively. The runtime impact of an inconsistency is therefore not a fixed property of the Skill alone.
\ding{184} These cross-model differences are not explained by Skill exposure. Across 441 paired executions, agents read the Skill in 97.1\% of runs. 
Exposure rates are similarly high for Haiku, Qwen, and DeepSeek at 98\%, 93\%, and 100\%, respectively.
\ding{185} The models diverge primarily after contact. Haiku, Qwen, and DeepSeek follow the inconsistency into failure in 30\%, 24\%, and 12\% of runs, respectively, while DeepSeek recovers in 56\% of exposed runs. The key distinction is therefore not whether an agent encounters the inconsistency, but how it responds after exposure.

Taken together, these findings show that static inconsistency count is a poor proxy for runtime risk. Runtime impact depends not only on whether an inconsistency exists, but also on whether it enters execution and how the agent responds after contact. Skill quality assessment should therefore move beyond counting static findings toward distinguishing their actual runtime risks.

\noindent\textbf{Contributions.}
In summary, this work makes the following contributions:

\noindent $\bullet$ \textbf{A taxonomy and large-scale empirical characterization of Skill contract inconsistencies.}
    We develop a taxonomy spanning three artifact boundaries and systematically analyze 1,182 manually confirmed inconsistencies across 575 real-world and benchmark Skills.

\noindent $\bullet$\textbf{A controlled evaluation of the runtime impact of Skill inconsistencies.}
    We construct 49 real-world tasks and quantify the effects of inconsistencies on task effectiveness and execution efficiency across three models under \textsc{NoSkill}, \textsc{OriginalSkill}, and \textsc{FixedSkill} conditions.

\noindent $\bullet$ \textbf{An execution-level account of how static inconsistencies translate into runtime impact.}
Through trajectory analysis, we trace inconsistencies from exposure to outcome and show that cross-model differences arise from post-exposure behavior rather than Skill exposure itself.

\section{Background and Motivation}

\subsection{Agent Skills}

Agent Skills are reusable execution units that provide LLM agents with structured procedural knowledge, tool usage logic, and domain-specific task workflows. Unlike ordinary prompts that provide guidance for a single request, a Skill packages multiple interdependent artifacts into a self-contained unit that an agent can discover, load, and execute.
A typical Skill is organized as a directory centered on a \texttt{SKILL.md} file. Its YAML frontmatter specifies metadata such as the Skill name, functionality, applicable scenarios, and dependencies, while the body describes execution steps, constraints, and expected outcomes. A Skill may also include scripts, configuration files, static resources, and other supporting artifacts.

\begin{figure}[t]
  \centering
  \includegraphics[width=0.8\linewidth]{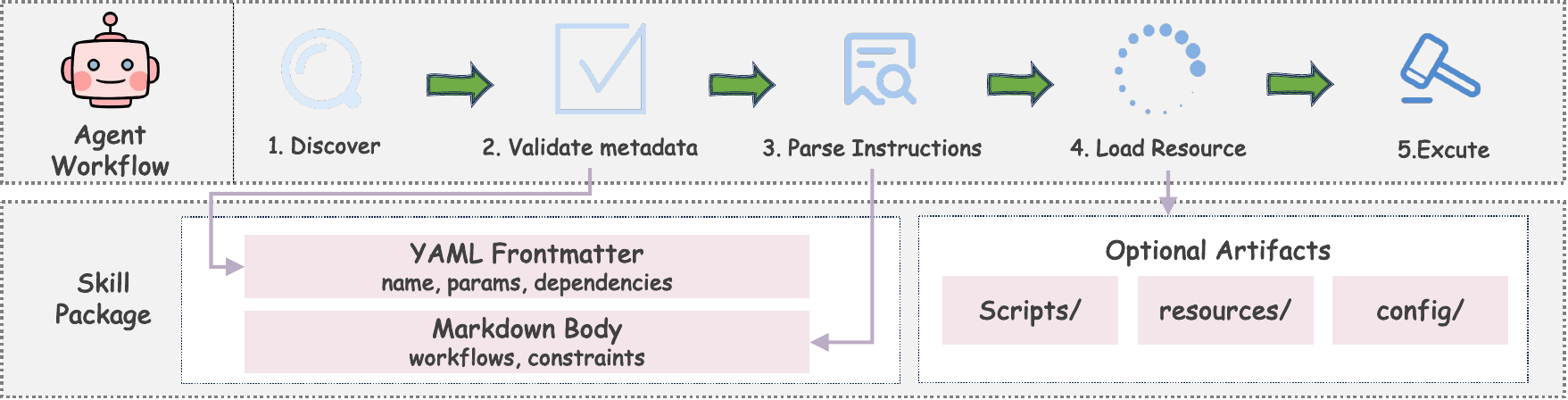}
  \caption{The workflow of using Skills.}
  \label{fig:workflowOfUseSkills}
\end{figure}

As shown in Figure~\ref{fig:workflowOfUseSkills}, an agent first identifies candidate Skills for a given task and determines their applicability from the associated metadata. It then reads the relevant instructions and resources, executes the prescribed workflow or code, and completes the task according to the requirements stated by the Skill.
A Skill is therefore more than a single textual prompt. Its documentation, code, resources, and environment requirements collectively define its expected behavior, while these artifacts may evolve independently. Reliable Skill execution depends not only on the correctness of each artifact, but also on whether they remain consistent about the same behavioral requirements.

\subsection{A Motivating Example}

From the perspective of an executing agent, the documentation, metadata, code, and resources packaged within a Skill collectively form an implicit \emph{Skill contract}. This contract specifies the behavioral commitments made by the Skill, including its functionality, inputs and outputs, execution procedure, dependencies, and runtime requirements.

A Skill contract combines natural language, semi-structured configuration, and executable code. We define a \emph{contract inconsistency} as a case in which two or more artifacts make incompatible claims about the same contract property. Such inconsistencies may arise between environment requirements and actual dependencies, across documentation artifacts, or between documentation and code.
However, the static presence of an inconsistency does not imply runtime failure. File inspection can establish that conflicting claims exist, but it cannot determine whether the inconsistency will enter the agent's execution path or what effect it will eventually have.

\begin{figure*}[t]
    \centering
    \includegraphics[width=\textwidth]{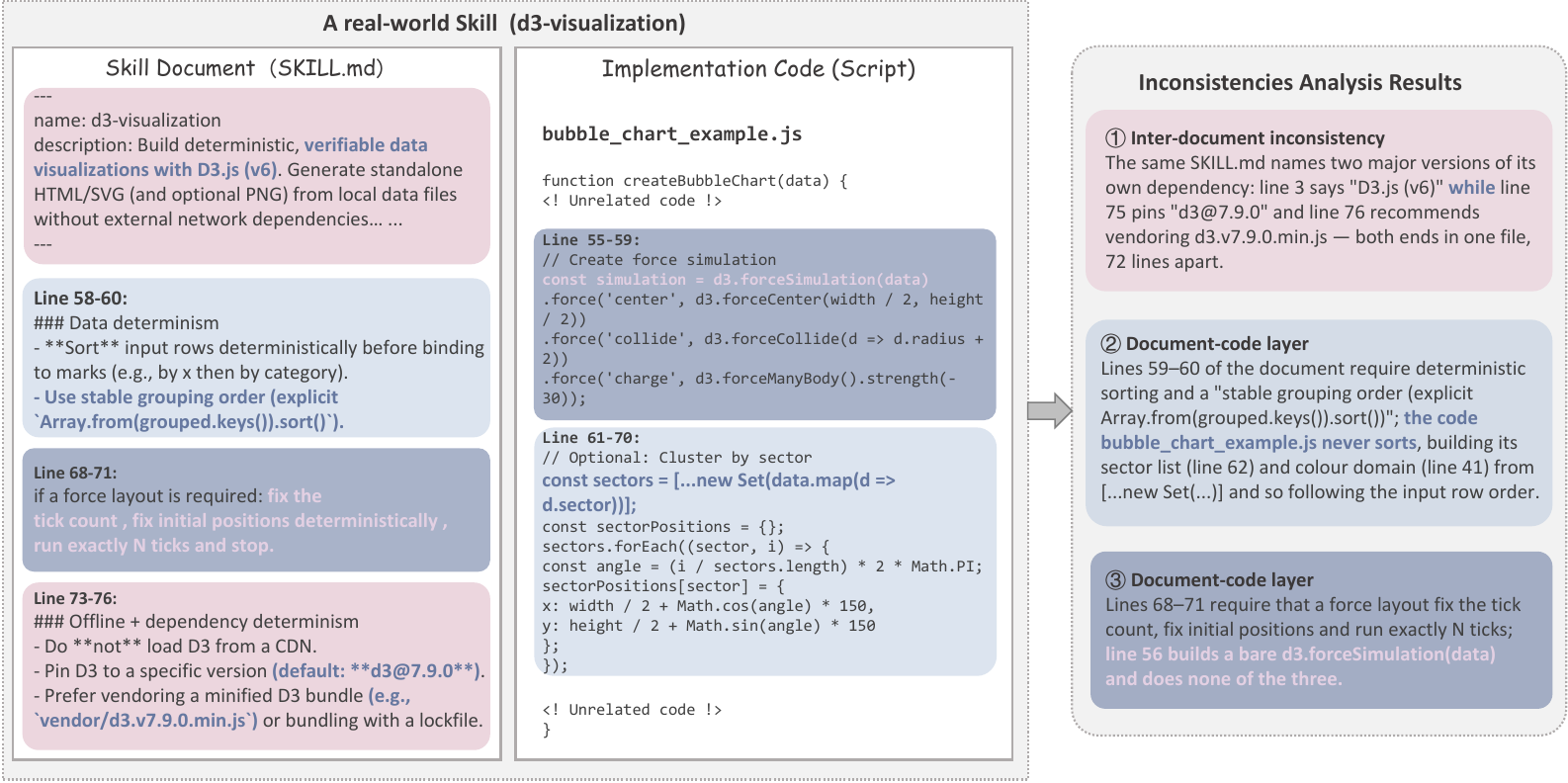}
\caption{A real-world Skill. \textcircled{1} marks an Inter Document inconsistency, while \textcircled{2} and \textcircled{3} mark Document Code inconsistencies.}
\label{fig:moti}
\end{figure*}

To illustrate this problem, we use the real-world \texttt{d3-visualization} Skill. Figure~\ref{fig:moti} shows three inconsistencies that can be identified through static inspection.

\noindent \ding{182} \textbf{Document-internal version conflict.}
One part of the documentation states that the Skill uses D3.js v6, while another specifies \texttt{d3@7.9.0} as the default dependency. These statements make incompatible claims about the same dependency version and therefore form an \emph{Inter Document} inconsistency.

\noindent \ding{183} \textbf{Document Code inconsistency in sorting behavior.}
The documentation requires deterministic sorting before binding input rows to visual elements.
However, \texttt{bubble\_chart\_example.js} directly constructs a \texttt{Set} from the input values without performing the required sorting. The resulting element order may therefore depend on the input order.

\noindent \ding{184} \textbf{Document Code inconsistency in force layout determinism.}
The documentation requires a fixed number of iterations, deterministic initial positions, and termination after the specified number of iterations. The corresponding code neither fixes the initial randomness nor enforces the specified iteration count. It therefore cannot guarantee the deterministic behavior described by the documentation.

This example leads to three observations.

\noindent\textbf{Observation 1. Skill contract inconsistencies are not limited to documentation and code.}
The first inconsistency exists entirely within the documentation and does not involve implementation behavior. This shows that Skill inconsistencies extend beyond conventional specification implementation mismatches. Understanding consistency problems in Skills therefore requires examining conflicts across multiple artifact boundaries.

\noindent\textbf{Observation 2. Static detectability does not reveal runtime impact.}
All three inconsistencies can be detected before execution, yet static evidence cannot determine whether they affect agent execution. It remains unclear whether such inconsistencies reduce task effectiveness, increase execution cost, or affect agents with different capabilities differently. Detecting an inconsistency therefore establishes only that a potential problem exists, not how much it matters at runtime.

\noindent\textbf{Observation 3. Final outcomes do not explain how runtime impact arises.}
Even when an inconsistency is associated with a change in task outcome, success or failure does not explain how that effect arises. We do not know whether the inconsistency enters the agent's execution path or how the agent responds after encountering the content. Understanding this process requires tracing how the inconsistency propagates through execution and how the agent responds to it.

Together, these observations reveal three open questions. We need to understand what inconsistencies exist in real Skills, how much they affect agent execution, and how those effects arise. We therefore ask the following Research Questions (RQs):

\noindent $\bullet$ \textbf{RQ1 (Prevalence and Characteristics).}
How prevalent are contract inconsistencies in Agent Skills, and what forms do they take?

\noindent $\bullet$ \textbf{RQ2 (Runtime Impact).}
To what extent do Skill contract inconsistencies affect the task effectiveness and execution efficiency of agents with different capabilities?

\noindent$\bullet$ \textbf{RQ3 (Impact Mechanisms).}
When and why do Skill contract inconsistencies affect agent execution?
\section{Study Methodology}
\begin{figure}[t]
  \centering
\includegraphics[width=0.9\linewidth]{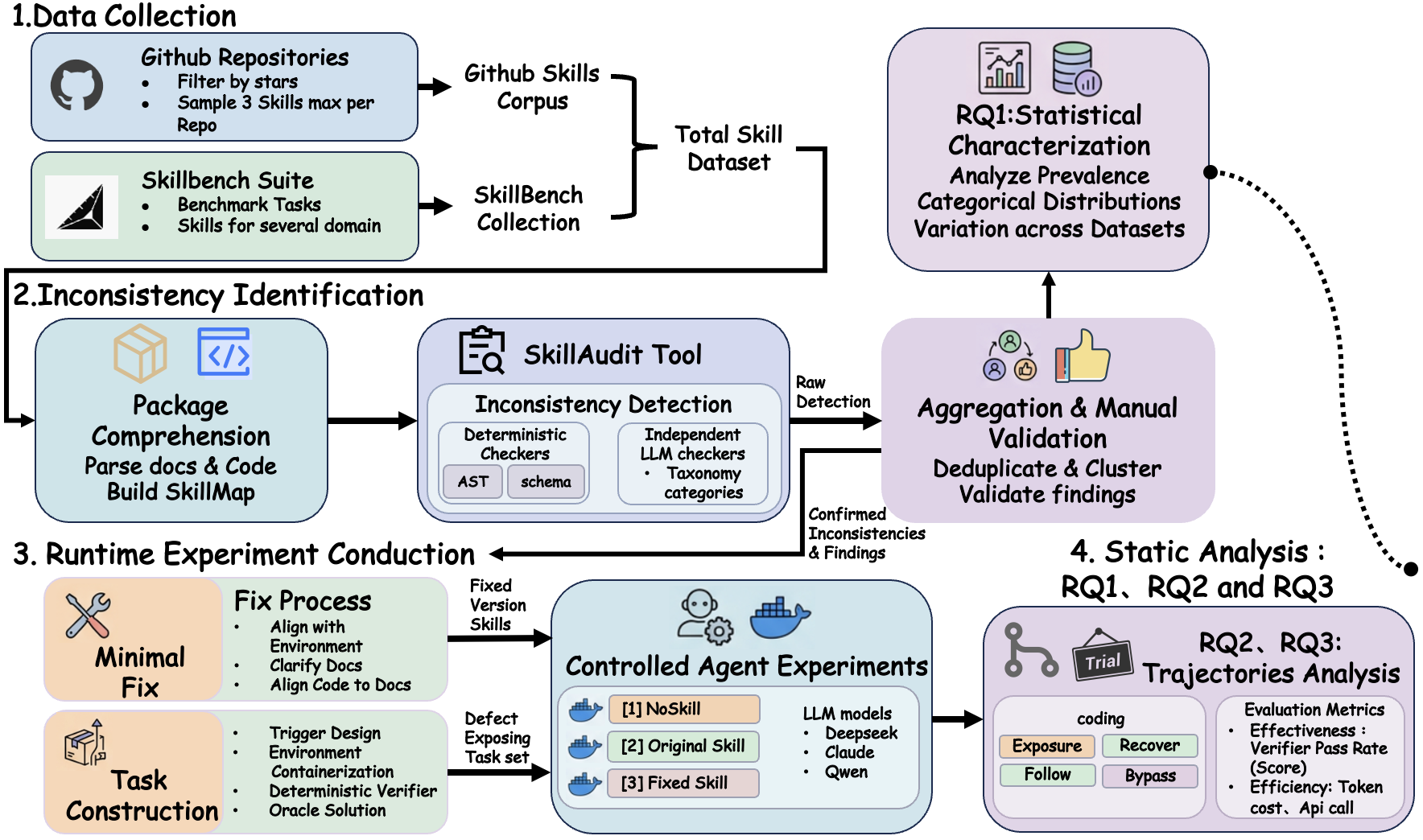}
  \caption{The Study pipeline of the paper.}
  \label{fig:pipelineOfStudy}
\end{figure}

Figure~\ref{fig:pipelineOfStudy} provides an end-to-end overview of our study pipeline. We first build a Skill dataset from GitHub and SkillBench and identify contract inconsistencies through taxonomy-guided SkillAudit and manual validation. RQ1 characterizes the prevalence, types, and distribution of these confirmed findings. We then transform selected real-world inconsistencies into triggerable and verifiable runtime tasks and construct minimally repaired counterparts. RQ2 evaluates their effects on task effectiveness and execution efficiency under \textsc{NoSkill}, \textsc{OriginalSkill}, and \textsc{FixedSkill} conditions. Finally, RQ3 analyzes paired execution trajectories to determine how inconsistencies enter execution and how agents respond after exposure.

\subsection{Data Collection}

We construct our dataset from GitHub and SkillBench to cover both naturally evolved real-world Skills and manually designed benchmark Skills. The final dataset contains 575 Skills.

\begin{table}[!ht]
  \centering
  \caption{Category distribution of the SkillBench and Github datasets.}
  \label{tab:skill-category-distribution}

  \newcommand{\ncell}[2]{\makecell{#1\\[-1.5pt]\scriptsize #2\%}}

  \resizebox{0.95\linewidth}{!}{%
    \begin{tabular}{lcccccccc}
      \toprule
      Source
      & \makecell{Software\\Engineering}
      & \makecell{Media \& Content\\Production}
      & \makecell{Finance \&\\Economics}
      & Cybersecurity
      & \makecell{Office \&\\White Collar}
      & \makecell{Industrial \&\\Physical Systems}
      & \makecell{Natural\\Science}
      & \makecell{Mathematics \&\\Operations Research} \\
      \midrule

      \makecell[l]{SkillBench\\[-1.5pt]\scriptsize ($n{=}87$)}
      & \ncell{16}{18.4}
      & \ncell{5}{5.7}
      & \ncell{9}{10.3}
      & \ncell{7}{8.0}
      & \ncell{14}{16.1}
      & \ncell{14}{16.1}
      & \ncell{14}{16.1}
      & \ncell{8}{9.2} \\

      \addlinespace[2pt]

      \makecell[l]{Github\\[-1.5pt]\scriptsize ($n{=}488$)}
      & \ncell{395}{80.9}
      & \ncell{36}{7.4}
      & \ncell{18}{3.7}
      & \ncell{5}{1.0}
      & \ncell{24}{4.9}
      & \ncell{2}{0.4}
      & \ncell{6}{1.2}
      & \ncell{2}{0.4} \\

      \bottomrule
    \end{tabular}%
  }
\end{table}

\subsubsection{GitHub Skill Collection.}
We sample community-maintained Skills from open-source GitHub repositories, prioritizing repositories with high star counts as a proxy for community visibility and practical use.
To capture structural differences, we divide candidate Skills into two groups. \textit{Document-only Skills} contain only Markdown documentation, without executable scripts or supplementary resources. \textit{Resource-augmented Skills} additionally include internal artifacts such as scripts, configuration files, or other supporting resources.
To reduce repository-level bias, we sample at most three Skills from each repository. We inspect Skill metadata and functional descriptions during selection and avoid choosing highly overlapping Skills from the same repository. This process yields 488 GitHub Skills.

\subsubsection{SkillBench Collection.}
We further include 87 Skills from SkillBench~\cite{li2026skillsbench}. Unlike community Skills that evolve naturally on GitHub, SkillBench Skills are manually authored and paired with explicit task objectives and ground-truth specifications, providing a more controlled Skill corpus.
This difference provides a useful contrast for RQ1. The GitHub corpus reflects consistency problems that arise during real-world development and maintenance, whereas SkillBench is manually constructed and reviewed, and therefore represents a more controlled artifact-quality setting. Comparing the two allows us to examine whether contract inconsistencies are primarily artifacts of real-world maintenance noise or remain common even in curated Skills.

\subsubsection{Data Summary.}
Across both sources, we obtain 575 Skills, as shown in Table~\ref{tab:skill-category-distribution}. Following the taxonomy used by SkillBench~\cite{li2026skillsbench}, we categorize them into eight domains: Software Engineering, Media \& Content Production, Finance \& Economics, Cybersecurity, Office \& White Collar Work, Industrial \& Physical Systems, Natural Science, and Mathematics \& Operations Research.



\subsection{Skill Contract Inconsistency Identification}
RQ1 first requires a consistent definition of Skill contract inconsistency and a reliable way to identify such cases at scale. We therefore construct an inconsistency taxonomy, operationalize it through a taxonomy-guided static audit pipeline, and manually validate the resulting findings.

\begin{figure}[t]
  \centering
  \includegraphics[width=0.75\linewidth]{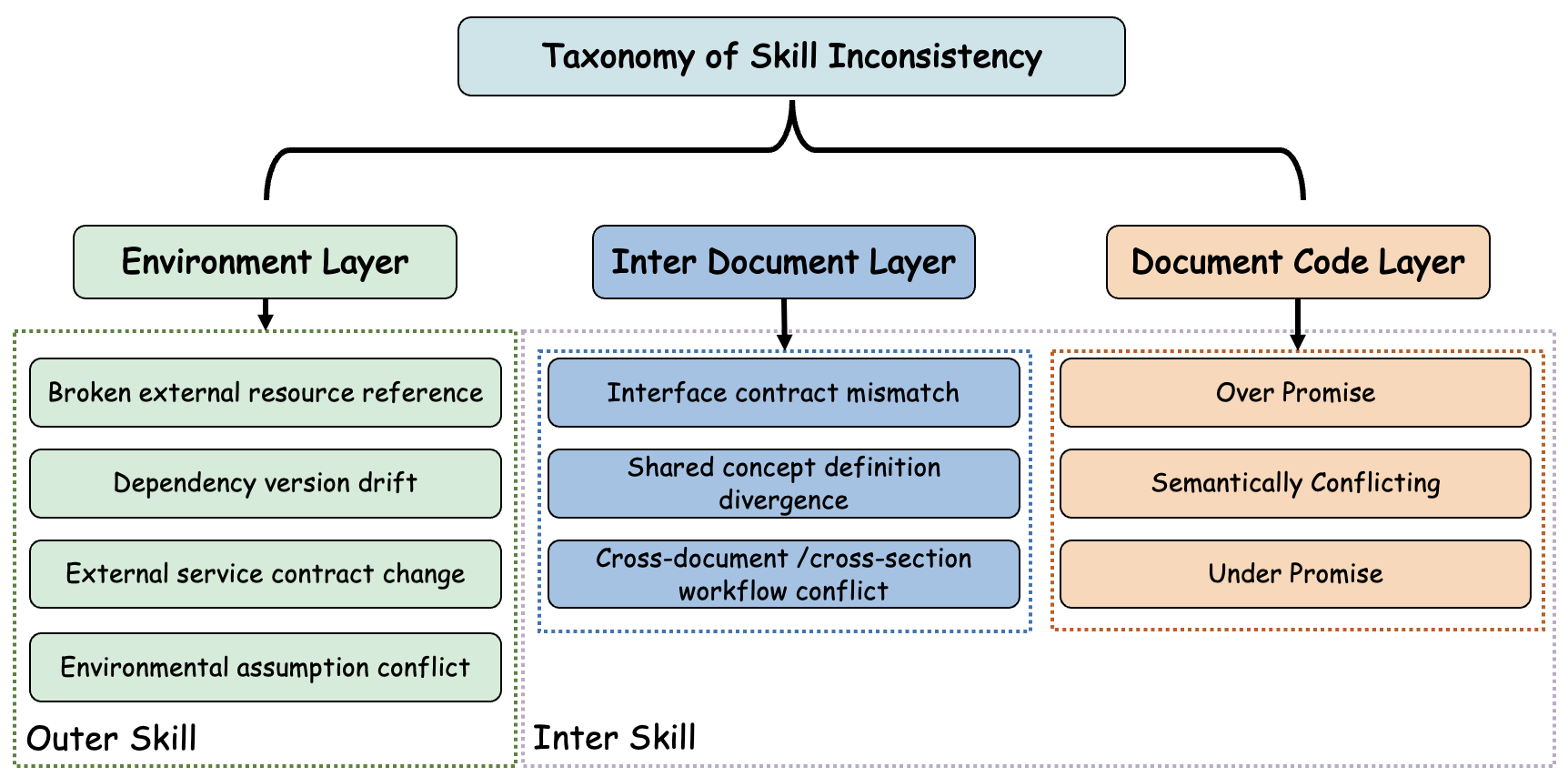}
  \caption{Taxonomy of skill‑level inconsistencies.}
  \label{fig:contract_taxonomy}
\end{figure}

\subsubsection{Taxonomy Of Skill Inconsistency}
\label{contractTaxonomy}
We construct the taxonomy through iterative qualitative analysis~\cite{raj2026model,guo2026malskillbench}. Starting from real-world Skill artifacts, including documentation, configuration, and source code, we manually inspect misalignment patterns and synthesize them with known issue patterns from tool-agent and API-contract literature~\cite{kiecker2026cascade,ma2026whenspec}. During iteration, we merge redundant categories, eliminate overlapping definitions, and refine the decision criteria for each class.
We first classify inconsistencies by the participants of the violated contract: $\bullet$ \textbf{Environmental layer:} Skill vs.\ runtime environment.
$\bullet$ \textbf{Inter-document layer:} conflicts among descriptive documents.
$\bullet$ \textbf{Document-code layer:} documentation vs.\ implementation.

Within the Document-code layer, we further classify inconsistencies by the direction of deviation:
 $\bullet$ \textbf{Over-promised (Doc $>$ Code).}
The documentation promises functionality, constraints, or behavior that the implementation does not provide.
 $\bullet$ \textbf{Semantically-conflicting (Doc $\neq$ Code).}
The documentation and implementation both specify the relevant behavior but contradict each other semantically.
 $\bullet$ \textbf{Under-promised (Doc $<$ Code).}
The implementation provides capabilities that are not documented.

Because this study focuses on contract violations that may affect agent execution, we primarily retain over-promised and semantically-conflicting cases. Under-promised cases are typically benign~\cite{he2026skill} and are suppressed by default. We also exclude failures caused purely by agent hallucination or external-service faults~\cite{tian2026aligning}, as these do not originate from inconsistency among Skill artifacts. The resulting taxonomy is shown in Figure~\ref{fig:contract_taxonomy}.




\subsubsection{Audit Tool}
To audit skill consistency, we design an tool named SkillAudit. SkillAudit operationalizes the proposed inconsistency taxonomy in \ref{contractTaxonomy}. It takes a skill package as input and outputs ranked findings mapped directly to canonical taxonomy classes. The tool performs purely static analysis without executing the target skill or requiring ground truth. Its pipeline comprises four sequential modules sharing a unified intermediate representation:

\noindent \ding{182} \textbf{Package Comprehension.} This module parses documentation and source code into a structured SkillMap. A scanner first validates the target directory (requiring at least one `SKILL.md`), filters out metadata, caches and binaries, and retains partial file contents for indexing. A two-round LLM agent then builds the SkillMap of components, sub-skills and relations: round one selects files for full reading; round two generates the structured map. The SkillMap is cached via content hashing to eliminate redundant model calls for unchanged packages.

\noindent \ding{183} \textbf{Inconsistency Detection.}
Two groups of checkers identify taxonomy-defined inconsistencies over the same evidence base. Four deterministic checkers perform AST-based code analysis, schema comparison, environment constraint checking, and execution-path URL validation. Five independent LLM checkers target different taxonomy categories and inspect semantic conflicts across artifacts. All detections are normalized to canonical taxonomy labels, while duplicate evidence from different checkers is retained for later aggregation.

\noindent \ding{184} \textbf{Adjudication and Aggregation.} This module refines raw detections into structured findings. An adjudicator reviews LLM findings in batches, assigning confirmed or suppressed verdicts with adjusted severity and evidence strength. Rule-based suppression is tracked separately. The adjudicator uses a fail-open design: errors preserve original findings to avoid silent defect loss. Aggregation follows two stages: exact-key deduplication, then semantic clustering by class, location and evidence overlap. Each cluster forms a single root cause entry listing all contributing checkers.

\noindent \ding{185} \textbf{Reporting.} The final reporting module produces reports with inconsistencies, root cause statistics, evidence strength distribution, and token usage. Each finding includes a taxonomy label, line-numbered evidence, rationale and evidence strength. Evidence strength is structurally enforced: only verified findings may claim execution blocking. Deterministic checkers run standalone without credentials or network access, ensuring full reproducibility.

\subsubsection{Manual Validation}
To verify the final auditting results, two authors independently reviewed the audit reports of skills drawn from the full pipeline output of audit tool. with a third resolving disagreements. Of the 575 skills in the datasets, 1182 inconsistencies were confirmed finally.

 \subsection{Task Construction}
 \textsc{WildSkills} contains 49 dynamically verifiable task instances, all derived from GitHub Skills. Each task materializes a single re-verified inconsistency so that the experiment focuses on the effect of that inconsistency rather than the overall quality of the Skill.
Each task is constructed in five stages.

\noindent \ding{182} \textbf{Selection.}
Candidate inconsistencies are drawn from manually confirmed GitHub findings. Before inclusion, we re-check the current Skill files to ensure that the target inconsistency still exists and can be localized precisely.

\noindent \ding{183} \textbf{Trigger Design.}
We write a realistic user request whose most direct solution requires the artifact or code path associated with the target inconsistency. We also size the input so that bypassing the Skill and manually recomputing the result is more expensive than using the Skill, reducing opportunities to sidestep the defect.

\noindent \ding{184} \textbf{Environment.}
The shipped Skill, project workspace, and input data are packaged into a Docker environment. The Skill is mounted at runtime rather than baked into the image, allowing all experimental conditions to share the same environment while differing only in the provided Skill.

\noindent \ding{185} \textbf{Verifier.}
Each task includes a deterministic verifier that checks the produced artifact against the target contract using programmatic assertions only. No LLM-as-a-judge is used.

\noindent \ding{186} \textbf{Oracle.}
We construct a reference solution that passes all verifier checks, confirming that the task itself is solvable and that failure does not arise from an infeasible specification.

This process converts a static finding into a repeatable runtime question: what happens when the target inconsistency is placed on a relevant execution path?







\subsubsection{Counterfactual Skill Repair}
To estimate the runtime impact of each target inconsistency, we construct a minimally repaired \textsc{FixedSkill} for every \textsc{OriginalSkill}. The goal is not to redesign the Skill, but to remove the target inconsistency while changing as little else as possible.

The repair strategy depends on the violated contract boundary. For an \textit{environmental inconsistency}, we align the Skill with the actual runtime environment and pin the relevant requirement. For an \textit{inter-document inconsistency}, we identify an authoritative statement and unify conflicting descriptions while leaving the implementation unchanged. For a \textit{document-code inconsistency}, we first determine the intended contract and then modify the implementation to satisfy it. When the documented promise cannot be supported under the target environment, we instead narrow the documentation to match the feasible behavior.

All repairs follow two constraints. First, each edit is minimal and confined to the inconsistent element. Second, the repair must not leak the task solution or mirror the verifier.
Each \textsc{WildSkills} task therefore has a semantically matched \textsc{OriginalSkill} and \textsc{FixedSkill}, forming the basis for controlled comparison.


\subsection{Models and Experimental Conditions}

We evaluate three models: DeepSeek-V4-Flash, Claude Haiku 4.5, and Qwen3.6-35B-A3B. Their performance on benchmarks and on our \textsc{NoSkill} baseline allows us to examine whether inconsistency impact varies across models, as shown in Table~\ref{tab:models}.
We emphasize within-model comparisons. The Qwen deployment uses 4-bit quantization, while DeepSeek-V4-Flash runs in its default non-thinking configuration. Absolute cross-model differences should therefore be interpreted with caution.
Each task is packaged as a fixed experimental unit containing the user instruction, workspace, Skill directory, deterministic verifier, and oracle solution, and is executed in an isolated Docker container. The agent runs as an unprivileged user, and the Skill directory is injected only at runtime.

\begin{table}[!ht]
  \centering
  \caption{Models and baseline performance.}
  \label{tab:models}

  \small
  \setlength{\tabcolsep}{4pt}

  \resizebox{0.55\linewidth}{!}{%
    \begin{tabular}{lccc}
      \toprule
      Model & Active / total params & SWE-bench Verified & Ours (\textsc{NoSkill}) \\
      \midrule

      DeepSeek-V4-Flash
        \citep{deepseek-v4,continuum-v4flash,macaron-v4bench}
        & 13B / 284B
        & \textbf{79.0}
        & \textbf{0.689} \\

      Claude Haiku 4.5
        \citep{anthropic-haiku45}
        & ---
        & 73.3
        & 0.534 \\

      Qwen3.6-35B-A3B
        \citep{qwen36-35b,omlx-qwen36}
        & \textbf{3B} / 35B
        & 73.4
        & 0.498 \\

      \bottomrule
    \end{tabular}%
  }
\end{table}

We evaluate three conditions:
\noindent $\bullet$ \textbf{\textsc{NoSkill}.}
No Skill artifact is provided.
\noindent $\bullet$ \textbf{\textsc{OriginalSkill}.}
The Skill containing the inconsistency is provided.
\noindent $\bullet$ \textbf{\textsc{FixedSkill}.}
The repaired Skill with the target inconsistency removed is provided.
The paired comparison between \textsc{OriginalSkill} and \textsc{FixedSkill} estimates the runtime effect of the target inconsistency, while \textsc{NoSkill} measures the effect of using the Skill relative to having no Skill.
All runs use the same execution harness and a uniform budget covering generation requests, cumulative tokens, and output tokens. Each task-condition combination is executed with three replicates to reduce stochastic variation.

\subsection{Metrics} \label{sec:metrics}

\noindent \textbf{Effectiveness.}
Our primary metric is verifier pass rate, defined as the fraction of deterministic checks satisfied by the produced artifact, ranging from 0 to 1. For each task-condition pair, we first average the three technical replicates and then perform task-level paired comparisons. The fix effect is defined as $\Delta = \textsc{Fixed} - \textsc{Original}$.
This yields 49 paired task-level observations per model.

\noindent \textbf{Efficiency.}
We record cumulative input and output tokens, API calls, agent turns, and wall-clock time. Runs that reach the per-run budget limit are treated as truncated and reported separately. For non-truncated runs, we report median execution costs and additionally use tokens per score point to capture the relationship between computational cost and task effectiveness.

\subsection{Execution Trajectory Collection and Coding}
We analyse the 441 paired trials used by RQ2 (49 tasks $\times$ 3 models $\times$ trials
r1--r3), each contrasting \textsc{OriginalSkill} against \textsc{FixedSkill} on an identical
task, resolved under the same cell-selection rule as RQ2 (current task-package image, latest
study among duplicates).

\begin{table}[!ht]
  \centering
  \caption{Trajectory coding scheme.}
  \label{tab:coding}

  \small
  \setlength{\tabcolsep}{3pt}
  \renewcommand{\arraystretch}{1.05}

  \resizebox{0.97\linewidth}{!}{%
    \begin{tabular}{@{}p{0.1\linewidth}p{1.2\linewidth}@{}}
      \toprule
      Code & Definition \\
      \midrule

      Exposed
        & Agent reads the skill, either from \texttt{skills/} or through the
          \texttt{Skill} tool. \\

      Unused
        & Agent never reads the skill and the inconsistency lies outside execution. \\

      Bypass
        & Agent is exposed but does not use the inconsistent element. No changed
          file is executed and no shipped-only form is authored. \\

      Follow
        & The inconsistent behaviour reaches the artifact. A tied verifier check
          fails in \textsc{OriginalSkill} and passes in \textsc{FixedSkill}. \\

      Recover
        & The inconsistent behaviour is triggered, but the tied checks still pass. \\

      BothFail
        & The tied check fails in both trials and the fix is not decisive. \\

      \bottomrule
    \end{tabular}%
  }
\end{table}

\begin{figure}[t]
\centering
\begin{subfigure}{0.45\linewidth}
  \includegraphics[width=\linewidth]{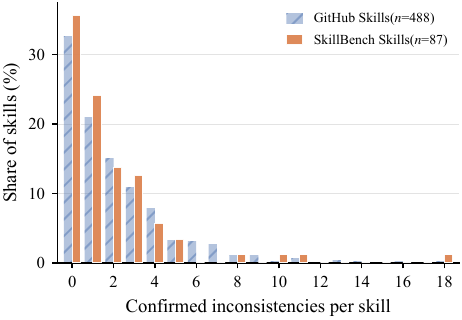}
  \caption{GitHub vs.\ SkillBench}
  \label{fig:dist_gh_skillbench}
\end{subfigure}
\hfill
\begin{subfigure}{0.45\linewidth}
  \includegraphics[width=\linewidth]{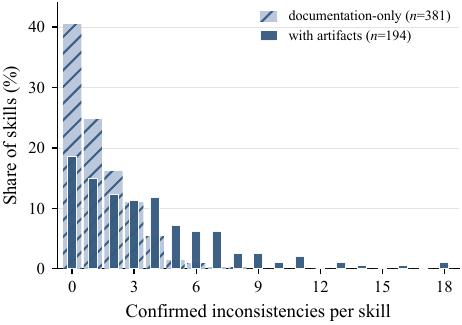}
  \caption{Documentation-only vs.\ with artifacts}
  \label{fig:dist_doc_artifact}
\end{subfigure}

\caption{Distribution of confirmed contract inconsistency counts per skill. The y-axis shows the proportion of skills in each bin of per-skill counts.}
\label{fig:inconsistency_distributions}
\end{figure}

For each task, the fix diff yields three signal types: the files the fix changed,
code-like tokens present only in the shipped skill (\emph{shipped-only anchors}), and tokens
present only in the fixed skill (\emph{fixed-only anchors}), together with the verifier
checks tied to the changed behaviour.
A deterministic coder walks each trajectory, separating what the agent \emph{authored} in
tool arguments from what tools \emph{returned} in file contents or command output, and records
skill contact, execution of a changed file, manifestation of shipped-only anchors,
per-check verifier failures, and turn-level timestamps.

Two decisions carry the analysis throughout.
First, \emph{contact is not use}: reading a file that contains an anchor is not adopting it;
only an authored shipped-only form, or execution of a changed file, counts as use.
Second, \emph{response is attributed by pairing}: a behaviour is attributed to the
inconsistency only when the \textsc{FixedSkill} arm of the same cell does not exhibit it,
which separates the inconsistency's contribution from a model's unconditional habits.
Table~\ref{tab:coding} gives the full operational definitions.

\section{RQ1: Characterizing Skill Contract Inconsistencies}

\subsection{Prevalence and Density}

Table \ref{tab:finding_stats} summarizes the prevalence of contract inconsistencies in both corpora,
reporting the skill population, the share of skills with at least one
inconsistency, and the total, mean, and maximum findings per skill. Among the
488 GitHub skills, 329 (67.4\%) contain at least one inconsistency. This corpus
yields 1,022 findings in total, averaging 2.09 per skill with a maximum of 18.
The SkillBench corpus shows a similar picture. 56 of its 87 tasks (64.4\%) contain
at least one confirmed inconsistency, with 160 findings in total, an average of
1.84 per skill, and the same maximum of 18.

\begin{table}[!ht]
  \centering
  \caption{Inconsistent findings by corpus and skill structure.}
  \label{tab:finding_stats}

  \small
  \setlength{\tabcolsep}{3pt}

  \resizebox{0.7\linewidth}{!}{%
    \begin{tabular}{llcccccc}
      \toprule
      Dataset & Category & \#Skills & $\geq$1 Inc. & Total Inc. & Mean & Max
        & Inc./kLine \\
      \midrule

      GitHub Skills
        & Doc-only     & 340 & 202 (59.4\%) & 421     & 1.24 & 7  & 6.17 \\
        & w/ artifacts & 148 & 127 (85.8\%) & 601     & 4.06 & 18 & 0.98 \\
        & All          & 488 & 329 (67.4\%) & 1{,}022 & 2.09 & 18 & 1.50 \\

      \midrule

      SkillBench Skills
        & Doc-only     & 41 & 25 (61.0\%) & 59  & 1.44 & 8  & 2.05 \\
        & w/ artifacts & 46 & 31 (67.4\%) & 101 & 2.20 & 18 & 0.91 \\
        & All          & 87 & 56 (64.4\%) & 160 & 1.84 & 18 & 1.14 \\

      \bottomrule
    \end{tabular}%
  }
\vspace{1pt}
\begin{minipage}{\linewidth}
  \footnotesize\centering
  \emph{Doc-only}: documentation-only skills. \quad
  \emph{Inc./kLine}: inconsistencies per thousand lines.
\end{minipage}
\end{table}

Figure~\ref{fig:inconsistency_distributions}(a) shows the per-skill
distribution of confirmed findings. Both corpora are heavily right-skewed, with
most skills carrying few inconsistencies and high-count cases being rare.
Although the two distributions share the same maximum of 18, GitHub skills show
a higher proportion of cases with large counts.
Figure~\ref{fig:inconsistency_distributions}(b) splits the skills by whether
implementation artifacts exist. Skills with available artifacts have fewer
zero-inconsistency cases and a heavier right tail. Concrete implementations
thus expose more mismatches between documented behavior and actual behavior.

\begin{findingbox}
\textbf{Finding 1.} Contract inconsistencies are widespread in real skills.
About two thirds of the skills in each corpus carry at least one, and most
affected skills contain only a few. Skills with implementation artifacts
accumulate more findings per skill, but doc-only skills show a higher per-line
rate because they are small and consist almost entirely of contract text.
Structural type alone is therefore not a reliable proxy for inconsistency risk.
\end{findingbox}

\subsection{Types of Inconsistencies}



Table~\ref{tab:rq1-taxonomy-distribution} breaks down the confirmed findings by
type, following the taxonomy introduced earlier. In the SkillBench corpus,
document--code inconsistencies dominate and account for 56.2\% of all findings.
Within this layer, over-promises are the most prevalent type and make up 45.6\%
of the total. The most common contract violation is therefore a documentation
that promises behavior the implementation fails to deliver. Semantic conflicts
occur but are far less frequent, and under-promised cases are rare in the
verified samples. The GitHub corpus shows a different profile, with the three
artifact boundaries contributing roughly equal shares. 

\begin{table}[!ht]
  \centering
  \caption{Distribution of contract inconsistencies by taxonomy.}
  \label{tab:rq1-taxonomy-distribution}

  \footnotesize
  \setlength{\tabcolsep}{4pt}

  \resizebox{0.68\linewidth}{!}{%
    \begin{tabular}{llrr}
      \toprule
      Artifact boundary & Inconsistency type & GitHub Skills & SkillBench Skills \\
      \midrule

      \multirow{4}*{Document--Code}
        & over\_promise                  & 250 (24.5\%) & 73 (45.6\%) \\
        & semantically\_conflicting      & 93 (9.1\%)   & 16 (10.0\%) \\
        & under\_promise                 & 1 (0.1\%)    & 1 (0.6\%) \\
      \cmidrule(lr){2-4}
        & subtotal                       & 344 (33.7\%) & 90 (56.2\%) \\

      \addlinespace[3pt]

      \multirow{4}*{Inter Document}
        & shared\_concept\_definition\_divergence
                                        & 162 (15.9\%) & 27 (16.9\%) \\
        & cross\_document\_workflow\_conflict
                                        & 81 (7.9\%)   & 2 (1.2\%) \\
        & interface\_contract\_mismatch & 69 (6.8\%)   & 16 (10.0\%) \\
      \cmidrule(lr){2-4}
        & subtotal                       & 312 (30.5\%) & 45 (28.1\%) \\

      \addlinespace[3pt]

      \multirow{5}*{Environment}
        & environmental\_assumption\_conflict
                                        & 274 (26.8\%) & 21 (13.1\%) \\
        & broken\_external\_resource\_reference
                                        & 44 (4.3\%)   & 3 (1.9\%) \\
        & external\_service\_contract\_change
                                        & 30 (2.9\%)   & 1 (0.6\%) \\
        & dependency\_version\_drift    & 18 (1.8\%)   & 0 (0.0\%) \\
      \cmidrule(lr){2-4}
        & subtotal                       & 366 (35.8\%) & 25 (15.6\%) \\

      \midrule
      \multicolumn{2}{l}{Total}
        & 1{,}022 (100.0\%) & 160 (100.0\%) \\
      \bottomrule
    \end{tabular}%
  }

  \vspace{1pt}
\end{table}

\begin{findingbox}
\textbf{Finding 2.} The two corpora differ in where inconsistencies occur.
GitHub skills split roughly evenly across the three artifact boundaries, while
the verified corpus is dominated by document-code inconsistencies. Most of
these are over\_promise, where the documentation promises behavior the
implementation fails to deliver. This asymmetry is the most common failure
mode and, as RQ3 will show, the one most likely to lead agents into failure.
\end{findingbox}

\subsection{Distribution across skill domains}
Table \ref{tab:rq1-domain} reports the share of skills with at least one confirmed inconsistency
within each application domain. Prevalence is high in both corpora, and the one
domain both sample densely agrees. Software engineering sits at 66\% for GitHub
and 69\% for SkillBench. The remaining domains are noisy; the two corpora largely
complement each other. The domains SkillBench samples densely are precisely those
GitHub undersamples, so no reliable per-domain contrast is possible outside
software engineering. Taken together, both corpora support the same qualitative
claim: no domain is immune, none is saturated, and inconsistency is a property of
skill contracts in general.

\begin{table}[!ht]
  \centering
\caption{Inconsistency prevalence across domains. }
  \label{tab:rq1-domain}

  \small
  \setlength{\tabcolsep}{3pt}

  \resizebox{0.8\linewidth}{!}{%
    \begin{tabular}{@{}llccccccccc@{}}
      \toprule
      Corpus & Metric
        & Software Engineering
        & Media
        & Office
        & Finance
        & Natural Science
        & Cybersecurity
        & Industrial
        & Math/Formal
        & All \\
      \midrule

      \multirow{2}{*}{GitHub Skills}
        & $n$    & 395 & 36 & 24 & 18 & 6 & 5 & 2 & 2 & 488 \\
        & Rate   & 66\% & 75\% & 71\% & 83\% & 50\% & 80\% & 100\% & 0\% & 67\% \\

      \midrule

      \multirow{2}{*}{SkillBench Skills}
        & $n$    & 16 & 5 & 14 & 9 & 14 & 7 & 14 & 8 & 87 \\
        & Rate   & 69\% & 60\% & 50\% & 56\% & 79\% & 71\% & 57\% & 75\% & 64\% \\

      \bottomrule
    \end{tabular}%
  }
\end{table}

\begin{findingbox}
\textbf{Finding 3.} Inconsistency is a property of skill contracts in general.
No domain with sufficient coverage is immune, and none is saturated.
\end{findingbox}




\section{RQ2: Runtime Impact of Skill Inconsistencies}


\subsection{Effectiveness}




Table~\ref{tab:rq2-effectiveness} summarizes effectiveness across the three
skill conditions. Fixing the inconsistency raises the verifier pass rate for
all three models, and the ordering $\textsc{NoSkill} \le \textsc{Original} \le
\textsc{Fixed}$ holds in every condition. The size of the gain varies by model.
Claude Haiku 4.5 improves the most, rising from 0.741 to 0.877. Qwen3.6-35B
improves moderately, from 0.707 to 0.792. DeepSeek-Flash gains the least, from
0.870 to 0.893.

\begin{table}[!ht]
  \centering
  \caption{Effectiveness across skill conditions on GitHub tasks.}
  \label{tab:rq2-effectiveness}

  \small
  \setlength{\tabcolsep}{4pt}

  \resizebox{0.55\linewidth}{!}{%
    \begin{tabular}{lccccc}
      \toprule
      Model & NoSkill & Original & Fixed & $\Delta$
        & Tasks $\uparrow$\,/\,$=$\,/\,$\downarrow$ \\
      \midrule

      Claude Haiku 4.5
        & 0.534 & 0.741 & \textbf{0.877}
        & $+0.136$ & 30\,/\,10\,/\,9 \\

      Qwen3.6-35B
        & 0.498 & 0.707 & \textbf{0.792}
        & $+0.085$ & 26\,/\,9\,/\,14 \\

      DeepSeek-Flash
        & 0.689 & 0.870 & \textbf{0.893}
        & $+0.023$ & 13\,/\,30\,/\,6 \\

      \bottomrule
    \end{tabular}%
  }

  \vspace{1pt}
  \begin{minipage}{\linewidth}
    \footnotesize\centering
    Score is the verifier pass rate. $\Delta$ is Fixed minus Original.
  \end{minipage}
\end{table}

The task-level breakdown shows that each average gain is driven by a subset of
tasks. For Haiku, 30 of 49 tasks improve, 10 stay unchanged, and 9 regress,
and the other two models follow the same pattern with fewer improving tasks.
For DeepSeek, the small overall delta is not an artifact of aggregation, since
the shipped and fixed skills yield identical scores on 30 of 49 tasks. A
cross-model comparison adds two observations. The baseline benefit of having
any skill is comparable across models, at between $+0.18$ and $+0.21$ for all
three. The additional gain from fixing the inconsistency, by contrast, spans
an order of magnitude. Moreover, Qwen's score under the fixed skill remains
below Haiku's. Its smaller gain therefore reflects a limited ability to
translate a correct procedure into passing runs, not only lower sensitivity to
inconsistency.

\begin{findingbox}
\textbf{Finding 4.} Resolving skill contract inconsistencies improves
effectiveness for all three models, and the ordering
$\textsc{NoSkill} \le \textsc{Original} \le \textsc{Fixed}$ is preserved
throughout. The size of the gain is strongly model-dependent. The
highest-capability model is largely robust, the mid-capability model gains the
most, and the lowest-capability model is bounded by its own ceiling. The
effect of an inconsistency is therefore a joint property of the contract and
the executing model.
\end{findingbox}

\begin{findingbox}
\textbf{Finding 5.} Task-level effects are highly heterogeneous. The aggregate
gains are concentrated on a minority of tasks, so mean deltas understate the
cost of inconsistency on the tasks where it actually matters. Per-task
analysis is needed to avoid masking this variation.
\end{findingbox}

\subsection{Efficiency}

Table~\ref{tab:rq2-efficiency} summarizes efficiency. Fixing the
inconsistency does not uniformly reduce runtime cost. For Haiku and Qwen, the
fixed condition consumes slightly fewer median tokens and less wall-clock time
than the original condition. For DeepSeek-Flash the trend reverses. Fixing the
inconsistency raises its median token consumption by 55k and its wall-clock
time by 20 seconds, and at 316k tokens and 69 seconds the shipped skill is its
cheapest condition. Cost per unit of effectiveness tells a more consistent
story. The tokens spent per score point fall from 1.22M to 0.69M for Haiku,
from 0.96M to 0.63M for Qwen, and from 0.76M to 0.57M for DeepSeek across the
three conditions. For DeepSeek, however, the shipped skill remains its most
efficient single condition.

\begin{table}[!ht]
  \centering
  \caption{Efficiency across skill conditions on GitHub tasks.}
  \label{tab:rq2-efficiency}

  \small
  \setlength{\tabcolsep}{3.5pt}

  \resizebox{0.7\linewidth}{!}{%
    \begin{tabular}{llcccccc}
      \toprule
      Model & Condition & Tokens (med.) & Calls & Turns
        & Time (s) & Trunc. (\%) & Tokens/point \\
      \midrule

      \multirow{3}{*}{Claude Haiku 4.5}
        & NoSkill  & 472k & 19 & 21 & 128 & 30 & 1.22M \\
        & Original & 472k & 16 & 19 & 124 & 24 & 0.83M \\
        & Fixed    & 435k & 16 & 21 & 110 & 20 & 0.69M \\

      \midrule

      \multirow{3}{*}{Qwen3.6-35B}
        & NoSkill  & 335k & 18 & 24 & 580 & 19 & 0.96M \\
        & Original & 362k & 15 & 21 & 508 & 15 & 0.71M \\
        & Fixed    & 344k & 14 & 21 & 468 & 18 & 0.63M \\

      \midrule

      \multirow{3}{*}{DeepSeek-V4-Flash}
        & NoSkill  & 440k & 20 & 25 & 145 & 26 & 0.76M \\
        & Original & 316k & 13 & 20 &  69 & 19 & 0.52M \\
        & Fixed    & 418k & 15 & 21 &  83 & 24 & 0.57M \\

      \bottomrule
    \end{tabular}%
  }

  \vspace{1pt}
  \begin{minipage}{\linewidth}
    \footnotesize\centering
    Medians exclude truncated runs. Trunc. is the share of runs reaching the budget limit.
  \end{minipage}
\end{table}

Budget truncation follows overall cost levels rather than correctness. In
every model, both skill conditions truncate less often than \textsc{NoSkill}.
The truncation rate falls from 30\% to 24\% for Haiku, from 26\% to 19\% for
DeepSeek, and from 19\% to 15\% for Qwen when any skill is provided. The
no-skill condition also issues the most API calls and turns. This pattern is
consistent with agents that lack a provided procedure spending extra effort to
search for a solution strategy. Fixing the skill reduces truncation further
only for Haiku, from 24\% to 20\%. For DeepSeek and Qwen it rises slightly, to
24\% and 18\% respectively.

\begin{findingbox}
\textbf{Finding 6.} Fixing inconsistencies does not automatically reduce
computational cost. Cost per unit of effectiveness improves for all models,
but absolute cost can increase, and for DeepSeek-Flash the inconsistent skill
is most efficient. A flawed skill's runtime signature is exploratory
interaction, with more API calls and turns, rather than uniformly longer
reasoning steps.
\end{findingbox}

\section{RQ3: When and Why Do Inconsistencies Matter?}

\subsection{From Static Inconsistency to Runtime Exposure}

\begin{table}[!ht]
  \centering
  \caption{Skill exposure by model and access route.}
  \label{tab:exposure}

  \small
  \setlength{\tabcolsep}{4pt}

  \resizebox{0.52\linewidth}{!}{%
    \begin{tabular}{@{}lrrrrrr@{}}
      \toprule
      Model & $n$ & Exposed & Rate & Path only & Tool only & Both \\
      \midrule

      Claude Haiku 4.5
        & 147 & 144 & 98\%   & 75  & 21 & 48 \\

      Qwen3.6-35B
        & 147 & 137 & 93\%   & 120 & 1  & 16 \\

      DeepSeek-V4-Flash
        & 147 & 147 & 100\%  & 119 & 1  & 27 \\
      \midrule

      All
        & 441 & 428 & 97.1\% & 314 & 23 & 91 \\

      \bottomrule
    \end{tabular}%
  }

  \vspace{1pt}
  \begin{minipage}{\linewidth}
    \footnotesize\centering
    Path denotes file-based access, while Tool denotes access through the
    \texttt{Skill} tool. Routes may overlap.
  \end{minipage}
\end{table}

\noindent \textbf{Contact Is Near-Universal.} Across all 441 executions the agent
reads the skill in 97.1\% of runs, and DeepSeek reads it in every run
(Table~\ref{tab:exposure}). Only 13 runs are unexposed. Eight reflect model
habit, where neither arm reads the skill, and five inconsistency-induced
avoidance, where only \textsc{OriginalSkill} skips it. Two points guard the
measure. Haiku reaches the skill through the \texttt{Skill} tool more often than
the other two models, with 69 runs against 28 and 17. Counting only path-based
access would understate exposure by five points and Haiku's by fifteen.
No execution inspects the \texttt{skills/} directory without also reading a
file, so the measure is not inflated by directory listing.

\noindent \textbf{Exposure Cannot Explain the RQ2 Gaps.}
The three exposure rates of 100\%, 98\%, and 93\% are nearly identical and are
not ordered by the RQ2 fix gains. Qwen's small residual gap partly reflects
its habit of leaving skills unread rather than an effect of the inconsistency.
Whether an agent sees the inconsistency is essentially constant across models
and therefore cannot explain why the fix benefit ranges from $+0.136$ to
$+0.023$.

\begin{findingbox}
\textbf{Finding 7.} Exposure to the inconsistency is near-universal across
models. Whether an agent sees the inconsistency does not differentiate the
models.
\end{findingbox}

\subsection{How Agents Respond to Skills}
\label{sec:rq3-response}


\noindent \textbf{The Capability Gap Operates after Contact.}
The models are indistinguishable in exposure but sharply distinguishable in
what they do next (Table~\ref{tab:response}). Haiku follows the inconsistency
into failure in 30\% of its runs, Qwen in 24\%, and DeepSeek in only 12\%.
Recovery shows the reverse ordering. DeepSeek recovers in 56\% of its exposed
runs, while Haiku and Qwen recover in 27\% and 22\% respectively. The
decisive step is whether the agent adopts what it reads.

\begin{table}[!ht]
  \centering
  \caption{Response distribution by model.}
  \label{tab:response}

  \small
  \setlength{\tabcolsep}{4pt}

  \resizebox{0.6\linewidth}{!}{%
    \begin{tabular}{@{}lrrrrrrr@{}}
      \toprule
      Model
        & Recover $\rightarrow$ success
        & Follow $\rightarrow$ failure
        & BothFail
        & Used, no effect
        & Bypass
        & Unused
        & All \\
      \midrule

      Haiku
        & 40
        & \textbf{44}
        & 40
        & 9
        & 11
        & 3
        & 147 \\

      Qwen
        & 33
        & 35
        & 43
        & 11
        & 15
        & 10
        & 147 \\

      DeepSeek
        & \textbf{82}
        & 18
        & 20
        & 19
        & 8
        & 0
        & 147 \\

      \midrule
      All
        & 155
        & 97
        & 103
        & 39
        & 34
        & 13
        & 441 \\

      \bottomrule
    \end{tabular}%
  }
\end{table}

\noindent \textbf{Follow-into-Failure Orders the Models as RQ2 Does.}
The per-model counts of follow-into-failure, 44 for Haiku, 35 for Qwen, and
18 for DeepSeek, reproduce the RQ2 fix-gain ordering exactly. The aggregate
fix effect is therefore the count of executions in which the agent adopts the
inconsistency and delivers it to the artifact. This ordering is visible only
under the cell-selection rule of RQ2. Resolving duplicate cells by manifest
order yields 34 for Haiku and 35 for Qwen, which spuriously inverts the
ordering.

\begin{figure*}[t]
  \centering
  \includegraphics[width=0.9\textwidth]{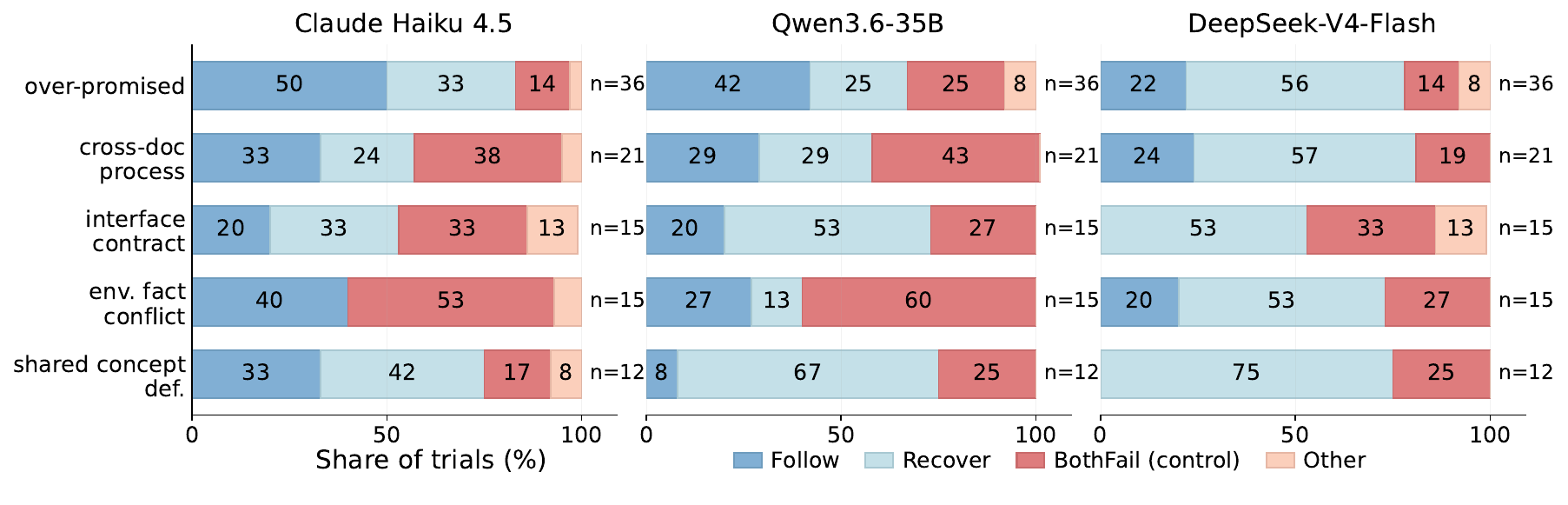}
  \caption{Response composition by inconsistency type and model.  \textsc{BothFail} denotes the control class.}
  \label{fig:response-type}
\end{figure*}

\noindent \textbf{Follow-into-Failure Is Concentrated in \texttt{over\_promise}.}
Figure~\ref{fig:response-type} disaggregates responses by model and
inconsistency type. The model gradient holds within every type and is steepest
in \texttt{over\_promised}, where the follow share ranges from 22\% for
DeepSeek to 50\% for Haiku. Of the 97 follow cases across all types, 41 arise
from \texttt{over\_promised}, where the documentation promises a capability
the implementation never provides. Conversely, \texttt{semantic\_conflict}
recovers in 13 of its 18 cases. A contradiction in meaning is noticed and
worked around far more often than a promised but absent capability.

\begin{findingbox}
\textbf{Finding 8.} Models diverge only after reading the skill.
Higher-capability models recover from inconsistencies far more often and
follow them into failure far less often. The per-model count of
follow-into-failure executions reproduces the ordering of the RQ2 fix gains
exactly. The delivery of the inconsistent artifact is therefore the mechanism
behind the score deltas.

\textbf{Finding 9.} Follow-into-failure is concentrated in
\texttt{over\_promise}. Documentation that promises a capability the
implementation never provides accounts for the largest share of follow cases,
while semantic conflicts are recovered most often. The type of inconsistency
is thus a strong predictor of whether the agent fails.
\end{findingbox}

\subsection{Runtime Consequences of Exposed Inconsistencies}

\label{sec:rq3-outcome}

\noindent \textbf{Fix Benefit Is Concentrated in the Follow Class.}
The fix's entire score benefit is concentrated in one cell of the matrix
(Table~\ref{tab:outcome}). Following the inconsistency into failure costs
$+0.366$ score points on average, and the fix recovers that loss. Every other
response class sits within $\pm 0.05$ of zero. Fixing a skill pays off exactly
when it stops the agent from delivering the inconsistency to the final
artifact. When a model does follow an inconsistency, the per-incident score
loss is nearly total, ranging from $+0.351$ to $+0.393$ by model. DeepSeek
suffers the same per-incident loss as Haiku but incurs 4.9$\times$ fewer such
incidents, which is why its RQ2 aggregate gain is small.

\begin{table}[!ht]
  \centering
  \caption{Outcome by response class.}
  \label{tab:outcome}

  \small
  \setlength{\tabcolsep}{3.5pt}

  \resizebox{0.55\linewidth}{!}{%
    \begin{tabular}{@{}lrrrrr@{}}
      \toprule
      Response & $n$ (fires) & $\Delta$score & net $\Delta$token
        & Footprint turns & Footprint (s) \\
      \midrule

      Follow $\rightarrow$ failure
        & 97 (87)   & $+0.366$ & $+34$k & 14.4 & 104 \\

      Recover $\rightarrow$ success
        & 155 (155) & $-0.038$ & $-57$k & 12.3 & 60 \\

      Bypass
        & 34 (4)    & $+0.097$ & $+2$k  & 3.8  & 40 \\

      BothFail (control)
        & 103 (87)  & $+0.030$ & $+7$k  & 19.6 & 169 \\

      Used, no effect
        & 39 (3)    & $-0.047$ & $-44$k & 11.3 & 24 \\

      Unused
        & 13 (3)    & $+0.127$ & $+4$k  & 16.0 & 88 \\

      \bottomrule
    \end{tabular}%
  }
\end{table}

\noindent \textbf{Recover Carries a Detour, Not a Net Compute Penalty.}
The 155 Recover runs spend 12.3 turns on the wrong path before correction,
which is 28\% of the run or about 116k tokens. This detour is not a net
compute penalty. The Fixed arm of the same pairs spends 57k tokens more than
the Original arm, because a more complete skill induces more work that offsets
the detour. Fixing buys correctness at no net compute saving. Footprint in
seconds is not comparable across models, since the local 4-bit Qwen deployment
is an order of magnitude slower than the API-hosted models. Turn counts are
the appropriate footprint measure.

\begin{table}[!ht]
  \centering
  \caption{Decomposition of the \textsc{BothFail} control class.}
  \label{tab:bothfail}

  \small
  \setlength{\tabcolsep}{5pt}

  \resizebox{0.45\linewidth}{!}{%
    \begin{tabular}{@{}lrrr@{}}
      \toprule
      Fix effect & $n$ & Share & Mean $\Delta$score \\
      \midrule

      Failure count reduced (partial fix)
        & 38 & 37\% & $+0.352$ \\

      Failure count unchanged
        & 37 & 36\% & $+0.011$ \\

      Failure count increased (adverse)
        & 28 & 27\% & $-0.380$ \\

      \midrule
      Total
        & 103 & 100\% & \\

      \bottomrule
    \end{tabular}%
  }
\end{table}

\noindent \textbf{BothFail Is Not a Uniform Task-Difficulty Floor.}
The 103 BothFail executions, 23\% of all paired runs, are not homogeneous
(Table~\ref{tab:bothfail}). In 38 of them the fix reduced the number of tied
checks that failed and recovered $+0.352$ of score, a genuine partial success.
In 37 the fix changed nothing, which is the true difficulty floor. In the
remaining 28 the fix made the outcome worse by $-0.380$ on average. These
adverse cases cluster in a handful of tasks, such as \texttt{promql-lint-gate},
\texttt{pdf-forms}, and \texttt{dingtalk-todo}, reproducing the negative-delta
tasks already visible in RQ2. Removing one inconsistency can still mislead
the agent by altering the execution path in an unexpected direction.

\begin{findingbox}
\textbf{Finding 10.} The score benefit of a fix is concentrated entirely in the runs that would
otherwise follow the inconsistency into failure; all other response classes
contribute nothing on net. Recovery preserves correctness but leaves a detour
footprint, so the fixed condition can spend more tokens. A fix buys correctness,
not compute savings.

\textbf{Finding 11.} The \textsc{BothFail} control class is heterogeneous.
About a third of its runs are partial fixes, about a third are a genuine
task-difficulty floor, and about a third are cases where the fix worsens the
outcome. Treating the class as a neutral floor would understate the benefit
of fixing and conceal the risk that a targeted fix can backfire.
\end{findingbox}

\section{Related Work}

\subsection{Agent Skills and Skill Ecosystems}

Agent skills consist of instructions, scripts, and resources that agents dynamically load at inference time to improve task performance \cite{anthropic-agent-skills-news, anthropic-agent-skills-docs}. Recent research on agent skills generally covers benchmarking, authoring, quality assurance, and security \cite{destefanis2026gitskills}.

Skill benchmark research constructs task collections and interactive environments to evaluate task completion, assessing factors such as absolute pass-rate gains and degradation risks from inference-time skill injection \cite{li2026skillsbench, han2026swe, tank2026regression}. Concurrently, work on skill authoring investigates how to define, reuse, and compose skills in natural language or code, while exploring automated generation, tuning, and evolution to lower authoring costs \cite{huang2026raw, ding2026agent}. From a quality perspective, studies treat skill definitions such as SKILL.md as executable artifacts to identify defects (found in 91.8\% of 138,133 evaluated skills) \cite{zhang2026keeps} by combining static rules, semantic reasoning, and automated repair \cite{meng2026skillconsist}. Finally, security-focused research addresses harmful skills that systematically lower safety refusal rates \cite{guo2026malskillbench, jiang2026harmfulskillbench}. in this domain, static detection struggles against adversarial evasion, and although sandboxed dynamic analysis remains the most reliable option \cite{ji2026cloaktonate}, existing detection tools still lag behind the rapid pace of threat evolution \cite{guo2026malskillbench, jiang2026harmfulskillbench}.

Despite these advances across different facets of the agent skill ecosystem, it remains unclear how static inconsistencies in skill definitions surface during runtime. Specifically, existing research lacks a systematic analysis of how agents respond to these defects and how such responses translate into measurable runtime costs \cite{dong2026agent}.

\subsection{Documentation--Code / API Contract Consistency}

Traditional software engineering research demonstrates that documentation--code inconsistency is pervasive in open-source projects, leading to developer misuse, increased debugging effort, and software defects \cite{tan2007comment, wen2019large}. Existing detection approaches span natural language--code consistency checking \cite{tan2007comment}, LLM-based reverse test generation \cite{kiecker2026cascade}, and Ethereum API specification--implementation verification \cite{ma2026whenspec}. However, as AI agents increasingly serve as direct consumers of tools and APIs, the nature of this problem shifts. Agents rely heavily on textual contracts, modifying a tool's description alone can alter an agent's tool selection \cite{faghih2025toolpref, sigdel2026schema}. Audits of production APIs reveal that many behavioral constraints exist solely in unstructured prose rather than formal schemas, leading to silent failures that full-loop agents often fail to detect or resolve, ultimately yielding false-negative outputs \cite{li2026silentprobe, shi2026descriptioncode}.

At the same time, agents do not passively succumb to these errors. Enhanced grounding, active verification, and structural guardrails can facilitate recovery, though performance ultimately depends on model capacity, environmental feedback, and task structure \cite{naik2026guardianagentbench, tank2026regression}. In dynamic environments with interdependent or unreliable tools, benchmarks indicate that agents frequently suffer from retrieval saturation, overconfidence, skipped validation steps, and failure rationalization \cite{li2026complexmcp}. Consequently, research must explicitly account for agent behavior, recovery, and bypass strategies, re-evaluating how inconsistent specifications impact agent reasoning, execution, and overall system reliability.

\subsection{Agent Reliability and Tool Use}

Existing research on agent reliability is largely agent-centric. Prior work attributes reliability to reasoning and planning \cite{yao2022react}, tool selection and invocation \cite{qin2024toolllm, patil2024gorilla}, benchmarks \cite{jimenez2024swebench, liu2024agentbench, zhou2024webarena}, and post-failure recovery \cite{shinn2023reflexion}. Consequently, researchers pursue robustness through better models, prompts, training, or pipelines. Evaluation surveys separate reliability and safety \cite{mohammadi2025survey, ding2026agent} while focusing on dynamic behavior. Meanwhile, failure-mode studies \cite{roig2025fail} identify archetypes and attribute them to model behavior.

A growing body of work connects static artifacts to reliability. Schema-izing prose constraints reduces silent failures \cite{li2026silentprobe}, and agents show high sensitivity to textual tool contracts \cite{faghih2025toolpref}. Recent years also witness the emergence of static skill auditing \cite{wen2026semia}, skill-behavior attribution \cite{tank2026regression, dong2026agent}, and interface alignment \cite{ren2026skillalign}. However, static auditing only targets constraint compliance, and interface alignment addresses runtime presentation. Existing work consequently fails to explain how static quality attributes are exposed upon invocation or how they influence failure, recovery, and bypass. This paper positions itself at the intersection of skill artifact quality and agent behavior. We conduct a white-box static census at the interface layer and treat static artifact quality as an independently auditable variable alongside agent capability to explain agent reliability.

\section{Discussion}

\subsection{Takeaways}

\noindent\textbf{For Skill Authors.}
Our results show that the number of inconsistencies does not directly indicate runtime risk. Although roughly two thirds of Skills contain at least one inconsistency, substantial harm is concentrated in a small set of inconsistency types. In particular, \texttt{over\_promised} inconsistencies are more likely to be adopted by agents and propagate into failures. When comprehensive auditing is infeasible, authors should therefore prioritize the Document Code boundary, especially cases where documentation promises behavior not supported by the implementation. For conflicts within documentation, authors should identify and unify the authoritative description. For conflicts between documentation and implementation, they should first determine the intended contract and then align the relevant artifacts with it.

\noindent\textbf{For Skill Users.}
The impact of Skill inconsistencies depends on the capability of the model using the Skill. Weaker models are more likely to continue along an incorrect path, so when using Skills that have not been thoroughly validated, users should prefer stronger models or Skills that have already been checked and repaired. Stronger models recover more often, but recovery still incurs additional execution cost. Therefore, in correctness-sensitive or latency-sensitive settings, users should prefer consistent and validated Skills even when using stronger models.

\noindent\textbf{For Future Skill Quality Tools.}
Reporting only the number of inconsistencies provides limited support for risk prioritization. Our results show that runtime risk also depends on inconsistency type, whether the relevant execution path is adopted, and the response capability of the target model. Future tools should therefore combine static findings with execution evidence and stratify findings by runtime risk rather than treating all inconsistencies equally. Beyond reporting how many issues are found, a more useful tool should identify which inconsistencies are more likely to enter execution and lead to follow, recover, or bypass behaviors.

\subsection{Threats to Validity}

\noindent\textbf{Task construction may increase inconsistency exposure.}
Our 49 runtime tasks were intentionally designed to trigger the target inconsistency and to reduce the chance that an agent bypasses the relevant path. This design helps measure the effect of an inconsistency once it enters execution, but it does not represent the natural exposure probability in real-world workloads. The high exposure rate observed in RQ3 should therefore be interpreted as a property of controlled defect-exposing tasks rather than as a general exposure rate in practical Skill use.

\noindent\textbf{Minimal fixes may still introduce side effects.}
RQ2 assumes that the main difference between \textsc{OriginalSkill} and \textsc{FixedSkill} is the targeted inconsistency. We reduce this risk through minimal edits and by preventing solution leakage and verifier mirroring. However, even a localized change may improve clarity or alter how an agent interprets the execution path. The difference between \textsc{FixedSkill} and \textsc{OriginalSkill} therefore primarily reflects the targeted inconsistency, but incidental effects of the repair cannot be fully excluded.

\noindent\textbf{Trajectory labels are operational approximations of agent behavior.}
The Exposed, Bypass, Follow, Recover, and BothFail labels in RQ3 are defined from observable execution traces, including file access, changed-file execution, artifact contents, and verifier outcomes. These labels capture external behavior rather than the agent's internal reasoning. For example, reading a file does not necessarily mean that the agent adopted its contents. Although paired attribution reduces confounding from model-specific behavior, these categories remain operational representations of the underlying runtime mechanisms.

\noindent\textbf{The runtime study covers only inconsistencies that can be operationalized as verifiable tasks.}
RQ1 identifies substantially more inconsistencies than the 49 cases studied in RQ2 and RQ3. Runtime experiments more readily include inconsistencies that can be triggered clearly, repaired minimally, and evaluated with deterministic verifiers. More ambiguous, environment-dependent, or difficult-to-verify inconsistencies may therefore be underrepresented. The runtime results should not be generalized directly to all inconsistency types identified in RQ1.

\section{Conclusion}
We study contract inconsistencies in Agent Skills through 575 Skills, 1,182
confirmed findings, and 49 runtime tasks across three models. Inconsistencies
are widespread, but their runtime impact varies substantially across models.
Although agents encounter the Skill in 97.1\% of executions, the key difference
is post-exposure behavior: models that follow inconsistent content into failure
benefit more from fixes, while models that recover are less affected. Static
inconsistency counts are thus poor proxies for runtime risk and motivate
runtime-aware Skill quality assessment.



\section*{Data Availability}
The artifact of this paper, containing all open source materials, can be
accessed via \url{https://anonymous.4open.science/r/SKill_Inconsistency-EC50/}.

\bibliographystyle{ACM-Reference-Format}
\bibliography{acmart}

@online{anthropic-agent-skills-news,
  author  = {{Anthropic}},
  title   = {Equipping agents for the real world with Agent Skills},
  year    = {2025},
  date    = {2025-10-16},
  url     ={https://www.anthropic.com/engineering/equipping-agents-for-the-real-world-with-agent-skills},
  urldate = {2026-09-22}
}

@online{anthropic-agent-skills-docs,
  author  = {{Anthropic}},
  title   = {Agent Skills},
  year    = {2026},
  url     = {https://platform.claude.com/docs/en/agents-and-tools/agent-skills/overview},
  urldate = {2026-09-22}
}

@inproceedings{wen2019large,
  title={A large-scale empirical study on code-comment inconsistencies},
  author={Wen, Fengcai and Nagy, Csaba and Bavota, Gabriele and Lanza, Michele},
  booktitle={2019 IEEE/ACM 27th International Conference on Program Comprehension (ICPC)},
  pages={53--64},
  year={2019},
  organization={IEEE}
}

@inproceedings{tan2007comment,
  title={/* icomment: Bugs or bad comments?*},
  author={Tan, Lin and Yuan, Ding and Krishna, Gopal and Zhou, Yuanyuan},
  booktitle={Proceedings of twenty-first ACM SIGOPS symposium on Operating systems principles},
  pages={145--158},
  year={2007}
}

@inproceedings{zhang2026keeps,
  title={What Keeps Agent Skills from Being Reusable? Evidence from 138K SKILL. md Files},
  author={Zhang, Chi and Liu, Yimin and Chen, Xinze and Ji, Ping},
  booktitle={First Workshop on Agent Skills},
  year={2026}
}

@article{naik2026guardianagentbench,
  title={GuardianAgentBench: Where Agents Fail and How to Guard Them},
  author={Naik, Vishal Ishwar and Xu, Chenyu and Dong, Donna and Hassan, Hussein and Pradhan, Abhishek and Mendelevitch, Ofer and Shafat, Tallat and Irshad, Humayun},
  journal={arXiv preprint arXiv:2607.20982},
  year={2026}
}

@article{jiang2026sok,
  title={SoK: Agentic Skills--Beyond Tool Use in LLM Agents},
  author={Jiang, Yanna and Li, Delong and Deng, Haiyu and Ma, Baihe and Wang, Xu and Wang, Qin and Yu, Guangsheng},
  journal={arXiv preprint arXiv:2602.20867},
  year={2026}
}

@article{destefanis2026gitskills,
  title={GitSkills: A Dataset of Agent Skills on GitHub},
  author={Destefanis, Giuseppe and Graziotin, Daniel and Vaccargiu, Matteo and Ortu, Marco},
  journal={arXiv preprint arXiv:2608.10906},
  year={2026}
}

@article{li2026skillsbench,
  title={SkillsBench: Benchmarking how well agent skills work across diverse tasks},
  author={Li, Xiangyi and Liu, Yimin and Chen, Wenbo and You, Bingran and Di, Zonglin and He, Yifeng and Zheng, Shenghan and Choe, Kyoung Whan and Sun, Jiankai and Wang, Shuyi and others},
  journal={arXiv preprint arXiv:2602.12670},
  year={2026}
}

@article{han2026swe,
  title={SWE-Skills-Bench: Do Agent Skills Actually Help in Real-World Software Engineering?},
  author={Han, Tingxu and Zhang, Yi and Song, Wei and Fang, Chunrong and Chen, Zhenyu and Sun, Youcheng and Hu, Lijie},
  journal={arXiv preprint arXiv:2603.15401},
  year={2026}
}

@article{meng2026skillconsist,
  title={SkillConsist: Detecting Inconsistencies in Agent Skills via Bidirectional Graph Alignment},
  author={Meng, Chaofan and Zheng, Yuhang and Zhou, Yingnan and Xu, Sihan},
  journal={arXiv preprint arXiv:2608.07639},
  year={2026}
}

@article{dong2026agent,
  title={Agent Skills Can Be Harmful: An Empirical Study of Skill-Induced Failures in LLM Agents},
  author={Dong, Gen and Gao, Yanjie and Li, Liqun and Xu, Tianyin and Hua, Yu and Yang, Fan},
  journal={arXiv preprint arXiv:2608.11888},
  year={2026}
}

@article{tank2026regression,
  title={The Regression Tax: Decomposing Why Skills Help and Hurt LLM Agents},
  author={Tank, Darshan and Nama, Baran},
  journal={arXiv preprint arXiv:2607.22520},
  year={2026}
}

@article{ding2026agent,
  title={Agent Skill Evaluation and Evolution: Frameworks and Benchmarks},
  author={Ding, Kexin and Zhou, Yang and Jin, Can and Tong, Feng and Zhou, Mu and Metaxas, Dimitris N},
  journal={arXiv preprint arXiv:2606.11435},
  year={2026}
}

@article{huang2026raw,
  title={From raw experience to skill consumption: A systematic study of model-generated agent skills},
  author={Huang, Zisu and Xu, Jingwen and Yang, Yifan and Gong, Ziyang and Yang, Qihao and Tian, Muzhao and Wang, Xiaohua and Lv, Changze and Gao, Xuemei and Dai, Qi and others},
  journal={arXiv preprint arXiv:2605.23899},
  year={2026}
}

@article{sigdel2026schema,
  title={Schema First Tool APIs for LLM Agents: A Controlled Study of Tool Misuse, Recovery, and Budgeted Performance},
  author={Sigdel, Akshey and Baral, Rista},
  journal={arXiv preprint arXiv:2603.13404},
  year={2026}
}

@article{he2026skill,
  title={Do Skill Descriptions Tell the Truth? Detecting Undisclosed Security Behaviors in Code-Backed LLM Skills},
  author={He, Wenhui and Li, Yue and Fu, Bang and Xing, Huan and Fan, Xing and Zhang, ZeHua and Niu, Baoning},
  journal={arXiv preprint arXiv:2605.12875},
  year={2026}
}

@article{shi2026descriptioncode,
  title={Description-Code Inconsistency in Real-world MCP Servers: Measurement, Detection, and Security Implications},
  author={Shi, Yutao and Zhang, Xiaohan and Zhang, Xiangjing and Shen, Xihua and Ouyang, Hui and Qiu, Huming and Zhang, Mi and Yang, Min},
  journal={arXiv preprint arXiv:2606.04769},
  year={2026}
}

@article{kiecker2026cascade,
  title={CASCADE: Detecting Inconsistencies between Code and Documentation with Automatic Test Generation},
  author={Kiecker, Tobias and Sparka, Jan Arne and Reuter, Martin and Ziegler, Albert and Grunske, Lars},
  journal={Proceedings of the ACM on Software Engineering},
  volume={3},
  number={FSE},
  pages={3816--3838},
  year={2026},
  publisher={ACM New York, NY, USA}
}

@article{raj2026model,
  title={Model or harness? an interaction-centric taxonomy for localizing agent failures},
  author={Raj, Harsh and Gupta, Vipul and Mahmoud, Anas and Dumitru, Razvan-Gabriel and Yi, Darvin and Sabharwal, Aakash and He, Yunzhong},
  journal={arXiv preprint arXiv:2607.28802},
  year={2026}
}

@article{guo2026malskillbench,
  title={MalSkillBench: A runtime-verified benchmark of malicious agent skills},
  author={Guo, Wenbo and Zeng, Wei and Liu, Chengwei and Jia, Xiaojun and Xu, Yijia and Tang, Lei and Fang, Yong and Liu, Yang},
  journal={arXiv preprint arXiv:2606.07131},
  year={2026}
}

@article{ma2026whenspec,
  title={When Specifications Meet Reality: Uncovering API Inconsistencies in Ethereum Infrastructure},
  author={Ma, Jie and He, Ningyu and Xi, Jinwen and Xing, Mingzhe and Liu, Liangxin and Luo, Jiushenzi and Fu, Xiaopeng and Wu, Chiachih and Wang, Haoyu and Gao, Ying and others},
  journal={Proceedings of the ACM on Programming Languages},
  volume={10},
  number={OOPSLA1},
  pages={540--570},
  year={2026},
  publisher={ACM New York, NY, USA}
}

@inproceedings{tian2026aligning,
  title={Aligning Requirement for Large Language Model's Code Generation},
  author={Tian, Zhao and Chen, Junjie},
  booktitle={Proceedings of the 2026 IEEE/ACM 48th International Conference on Software Engineering},
  pages={2951--2962},
  year={2026}
}

@article{jiang2026harmfulskillbench,
  title={Harmfulskillbench: How do harmful skills weaponize your agents?},
  author={Jiang, Yukun and Zhang, Yage and Backes, Michael and Shen, Xinyue and Zhang, Yang},
  journal={arXiv preprint arXiv:2604.15415},
  year={2026}
}

@article{ji2026cloaktonate,
  title={Cloak and Detonate: Scanner Evasion and Dynamic Detection of Agent Skill Malware},
  author={Ji, Zimo and Xu, Congying and Li, Zongjie and Gao, Yudong and Wei, Xin and Wang, Shuai and Cheung, Shing-Chi},
  journal={arXiv preprint arXiv:2607.02357},
  year={2026}
}

@article{li2026complexmcp,
  title={ComplexMCP: Evaluation of LLM agents in dynamic, interdependent, and large-scale tool sandbox},
  author={Li, Yuanyang and Yang, Xue and Wang, Longyue and Luo, Weihua and Chen, Hongyang},
  journal={arXiv preprint arXiv:2605.10787},
  year={2026}
}

@inproceedings{faghih2025toolpref,
  title={Tool preferences in agentic LLMs are unreliable},
  author={Faghih, Kazem and Wang, Wenxiao and Cheng, Yize and Bharti, Siddhant and Sriramanan, Gaurang and Balasubramanian, Sriram and Hosseini, Parsa and Feizi, Soheil},
  booktitle={Proceedings of the 2025 Conference on Empirical Methods in Natural Language Processing},
  pages={20965--20980},
  year={2025}
}

@article{li2026silentprobe,
  title={SilentProbe: Measuring Silent Failure in Production APIs Used as Agent Tools},
  author={Li, Zongrong and Ye, Shengkun and Guo, Feiyou and Dang, Zuoyou},
  journal={arXiv preprint arXiv:2609.00035},
  year={2026}
}

@article{wen2026semia,
  title={Semia: Auditing agent skills via constraint-guided representation synthesis},
  author={Wen, Hongbo and Li, Ying and Liu, Hanzhi and Shou, Chaofan and Chen, Yanju and Tian, Yuan and Feng, Yu},
  journal={arXiv preprint arXiv:2605.00314},
  year={2026}
}

@article{yao2022react,
  title={React: Synergizing reasoning and acting in language models},
  author={Yao, Shunyu and Zhao, Jeffrey and Yu, Dian and Du, Nan and Shafran, Izhak and Narasimhan, Karthik and Cao, Yuan},
  journal={arXiv preprint arXiv:2210.03629},
  year={2022}
}

@inproceedings{qin2024toolllm,
  title={Toolllm: Facilitating large language models to master 16000+ real-world apis},
  author={Qin, Yujia and Liang, Shihao and Ye, Yining and Zhu, Kunlun and Yan, Lan and Lu, Yaxi and Lin, Yankai and Cong, Xin and Tang, Xiangru and Qian, Bill and others},
  booktitle={International Conference on Learning Representations},
  volume={2024},
  pages={9695--9717},
  year={2024}
}

@article{patil2024gorilla,
  title={Gorilla: Large language model connected with massive apis},
  author={Patil, Shishir G and Zhang, Tianjun and Wang, Xin and Gonzalez, Joseph E},
  journal={Advances in Neural Information Processing Systems},
  volume={37},
  pages={126544--126565},
  year={2024}
}

@inproceedings{jimenez2024swebench,
  title={Swe-bench: Can language models resolve real-world github issues?},
  author={Jimenez, Carlos E and Yang, John and Wettig, Alexander and Yao, Shunyu and Pei, Kexin and Press, Ofir and Narasimhan, Karthik},
  booktitle={International Conference on Learning Representations},
  volume={2024},
  pages={54107--54157},
  year={2024}
}

@inproceedings{liu2024agentbench,
  title={Agentbench: Evaluating llms as agents},
  author={Liu, Xiao and Yu, Hao and Zhang, Hanchen and Xu, Yifan and Lei, Xuanyu and Lai, Hanyu and Gu, Yu and Ding, Hangliang and Men, Kaiwen and Yang, Kejuan and others},
  booktitle={International Conference on Learning Representations},
  volume={2024},
  pages={52989--53046},
  year={2024}
}

@inproceedings{zhou2024webarena,
  title={Webarena: A realistic web environment for building autonomous agents},
  author={Zhou, Shuyan and Xu, Frank F and Zhu, Hao and Zhou, Xuhui and Lo, Robert and Sridhar, Abishek and Cheng, Xianyi and Ou, Tianyue and Bisk, Yonatan and Fried, Daniel and others},
  booktitle={International Conference on Learning Representations},
  volume={2024},
  pages={15585--15606},
  year={2024}
}

@article{shinn2023reflexion,
  title={Reflexion: Language agents with verbal reinforcement learning},
  author={Shinn, Noah and Cassano, Federico and Gopinath, Ashwin and Narasimhan, Karthik and Yao, Shunyu},
  journal={Advances in neural information processing systems},
  volume={36},
  pages={8634--8652},
  year={2023}
}

@inproceedings{mohammadi2025survey,
  title={Evaluation and benchmarking of llm agents: A survey},
  author={Mohammadi, Mahmoud and Li, Yipeng and Lo, Jane and Yip, Wendy},
  booktitle={Proceedings of the 31st ACM SIGKDD Conference on Knowledge Discovery and Data Mining V. 2},
  pages={6129--6139},
  year={2025}
}

@article{roig2025fail,
  title={How do llms fail in agentic scenarios? a qualitative analysis of success and failure scenarios of various llms in agentic simulations},
  author={Roig, JV},
  journal={arXiv preprint arXiv:2512.07497},
  year={2025}
}

@article{ren2026skillalign,
  title={SkillAlign: Aligning Skill Interfaces for LLM-based Agents},
  author={Ren, Shuo and Kang, Xiaomian and Zhang, Jiajun},
  journal={arXiv preprint arXiv:2609.07255},
  year={2026}
}

@misc{anthropic-haiku45,
  title        = {Claude Haiku 4.5},
  author       = {{Anthropic}},
  year         = {2025},
  month        = oct,
  howpublished = {\url{https://www.anthropic.com/claude/haiku}},
}

@misc{qwen36-35b,
  title        = {{Qwen3.6-35B-A3B}},
  author       = {{Qwen Team}},
  year         = {2026},
  month        = apr,
  howpublished = {\url{https://huggingface.co/Qwen/Qwen3.6-35B-A3B}},
}

@article{deepseek-v4,
  title={Deepseek-V4: Towards highly efficient million-token context intelligence},
  author={{DeepSeek-AI} and Xu, Anyi and Lin, Bangcai and Xue, Bing and Wang, Bingxuan and Xu, Bingzheng and Wu, Bochao and Zhang, Bowei and Lin, Chaofan and Dong, Chen and Ling, Chenchen and others},
  journal={arXiv preprint arXiv:2606.19348},
  year={2026}
}

@misc{continuum-v4flash,
  title        = {Models \& Benchmarks --- {DeepSeek} {V4-Flash}},
  author       = {{Continuum Labs}},
  year         = {2026},
  howpublished = {\url{https://www.continuumlabs.ai/inference/models}},
}

@misc{macaron-v4bench,
  title        = {{DeepSeek} {V4} Benchmarks: {MMLU}, {HumanEval} \& {SWE-bench}},
  author       = {Hanks},
  year         = {2026},
  howpublished = {\url{https://macaron.im/blog/deepseek-v4-benchmarks}},
}

@misc{omlx-qwen36,
  title        = {{Qwen3.6-35B-A3B-evolved-mxbit} --- {MMLU} 78.7\%},
  author       = {{oMLX}},
  year         = {2026},
  month        = aug,
  howpublished = {\url{https://omlx.ai/benchmarks/intelligence/dr9anj1u}},
}

@article{wang2023voyager,
  title={Voyager: An open-ended embodied agent with large language models},
  author={Wang, Guanzhi and Xie, Yuqi and Jiang, Yunfan and Mandlekar, Ajay and Xiao, Chaowei and Zhu, Yuke and Fan, Linxi and Anandkumar, Anima},
  journal={arXiv preprint arXiv:2305.16291},
  year={2023}
}

@article{zhou2026comprehensive,
  title={A comprehensive survey on agent skills: Taxonomy, techniques, and applications},
  author={Zhou, Yingli and Shu, Wang and Su, Yaodong and Du, Wenchuan and Fang, Yixiang and Lin, Xuemin},
  journal={arXiv preprint arXiv:2605.07358},
  year={2026}
}





\end{document}